\documentclass[11pt]{article}

\usepackage[margin=1in]{geometry}
\usepackage{amsmath,amssymb}
\usepackage{booktabs}
\usepackage{graphicx}
\usepackage{microtype}
\usepackage[numbers,sort&compress]{natbib}
\usepackage[colorlinks=true,linkcolor=blue,citecolor=blue,urlcolor=blue,hyperfootnotes=false]{hyperref}

\title{When Do Geometric Algebra Layers Beat Scalarization?\\
A Controlled Study on SO(3)-Equivariant Vector Laws}

\author{Fabien Polly\\ \href{https://github.com/infinition/ga-vs-scalarization}{github.com/infinition/ga-vs-scalarization}}
\date{}

\begin{document}
\maketitle

\begin{abstract}
Compact networks built from Clifford algebra $\mathrm{Cl}(3,0)$ primitives
(grade-wise equivariant linear maps, geometric products, grade gates) are
exactly SO(3)-equivariant and learn synthetic 3D vector laws from few samples.
We ask whether the geometric algebra structure itself contributes anything
beyond exact equivariance. We compare against a minimal scalarization baseline:
invariant dot products fed to a small MLP that outputs coefficients on the
equivariant basis $\{v_i,\, v_i \times v_j\}$, which is also exactly
equivariant. On single-stage laws (rotation by axis-angle, cross product,
central force), scalarization matches or beats the $\mathrm{Cl}(3,0)$ network
at a fraction of the training cost, so the geometric algebra adds nothing
there. On compositional targets whose computation graph nests group
operations (apply $R_2 R_1$ to a point; map a local force through an
orientation, then take a torque), the $\mathrm{Cl}(3,0)$ network beats
scalarization by an order of magnitude in the low-data regime, reaching with
100 samples the accuracy the baseline achieves with 3000, and the gap survives
strengthening the baseline with the triple-product invariant and 17x more
parameters, external Vector Neurons and e3nn baselines, and a multiplicative
coefficient network. Ablations show the required network depth tracks the
rotation chain length, and scalarization falls below the constant predictor
on chains of four rotations. The advantage is not composition per se: on a
rotation-free nested cross product, which flattens into polynomial invariant
coefficients, scalarization wins by 24x. On the torque task the advantage is
confined to low data, with a crossover near a thousand samples. No tested
model, equivariant or not, extrapolates invariant
magnitudes: on radius and separation shifts every model is worse than a
constant predictor once errors are normalized. We conclude that geometric
algebra layers are not a general shortcut for low-data 3D learning, but become
useful precisely when the target composes group elements in depth.
\end{abstract}

\section{Introduction}

Many 3D learning problems require functions that commute with rotations: a
predicted force, torque, or displacement should rotate with the input frame.
Equivariant architectures encode this constraint in the parameterization, and
small-scale demonstrations routinely show large sample-efficiency gains over
unconstrained networks. Such demonstrations, however, usually conflate two
distinct effects: (a) restricting the hypothesis class to equivariant
functions, and (b) the specific parameterization of the proposed architecture.
Effect (a) is available from a trivial construction. By classical invariant
theory, every SO(3)-equivariant vector-valued function of vectors
$v_1,\dots,v_k$ can be written as
$\sum_j c_j(\text{invariants})\, b_j$, where the $b_j$ range over the inputs
and their pairwise cross products and the $c_j$ are functions of the SO(3)
invariants of the inputs: the pairwise dot products and, for orientation-sensitive
(chiral) targets, the scalar triple products $v_a \cdot (v_b \times v_c)$
~\citep{villar2021scalars}. Learning the $c_j$ with a small MLP
gives an exactly equivariant model that we call \emph{scalarization}. Our base
scalarization uses the dot-product invariants; the strengthened variant adds the
triple products, and the two tasks with three or more inputs (two-body, nested
cross) test whether that matters.

This paper asks a narrow question: does a small geometric algebra network
built from $\mathrm{Cl}(3,0)$ bilinear layers offer anything that
scalarization does not? We answer it with a controlled benchmark of six
synthetic vector laws, normalized metrics, tuning controls, and strengthened
baseline variants. Our contributions are negative and positive results of
equal importance:

\begin{enumerate}
\item On single-stage laws, those computable with one equivariant
interaction (rotation of a point, cross product, central force),
scalarization matches or beats the geometric algebra network, at 13 to 19x
lower training cost (Section~\ref{sec:results}, Finding 2).
\item On compositional laws, those whose computation graph nests group
operations (composition of two rotations, local-to-world torque), the
geometric algebra network wins by 3x to 16x in the low-data regime and under
angle shift, and the gap survives capacity and feature strengthening of the
baseline; on the torque task scalarization catches up beyond a thousand
samples (Finding 3, Figure~\ref{fig:curves}).
\item No tested model extrapolates invariant magnitudes. On radial
out-of-distribution splits, every model, equivariant or not, is worse than a
constant predictor once errors are normalized (Finding 4). Unnormalized MSE
hides this failure, which we argue is a reporting pitfall for equivariance
benchmarks.
\end{enumerate}

\section{Related work}
\label{sec:related}

Equivariant deep learning encodes group symmetry in the
architecture~\citep{cohen2016group,bronstein2021geometric}. For 3D rotations,
irreducible-representation approaches include Tensor Field
Networks~\citep{thomas2018tensor}, the e3nn framework~\citep{geiger2022e3nn},
and steerable networks~\citep{weiler2019general}; lighter-weight constructions
include Vector Neurons~\citep{deng2021vector} and E(n)-equivariant graph
networks~\citep{satorras2021egnn}. Equivariance is known to improve data
efficiency in the sciences~\citep{batzner2022nequip}. Geometric (Clifford)
algebra architectures embed inputs as multivectors and use the geometric
product as the core operation~\citep{brandstetter2022clifford,
ruhe2023clifford, brehmer2023geometric}. On the analysis side,
\citet{villar2021scalars} characterize equivariant functions through scalar
invariants, which directly motivates our scalarization baseline, and
\citet{dym2020universality} study the universality of rotation-equivariant
models. Closest to our baseline, \citet{domina2025spherical} express
equivariant targets as scalar functions multiplying a small tensor basis;
closest to our geometric model, GLGENN~\citep{filimoshina2025glgenn} builds
parameter-light Clifford equivariant networks. Unlike GATr, which scales a
projective geometric algebra transformer to large scenes, we work at the
opposite end: a minimal controlled benchmark isolating what the algebra
contributes. Our contribution is not a new architecture: it is a controlled
delimitation of when the geometric product parameterization pays relative to
the scalarization construction that theory already provides.

\section{Models}
\label{sec:models}

All models map 6 or 9 input dimensions (two or three 3D vectors) to one 3D
vector and are trained identically: Adam, full batch, 200 epochs, learning
rate $5\times 10^{-3}$, mean squared error. A grid over learning rate and
epochs (Section~\ref{sec:controls}) confirms that no conclusion is an
artifact of this choice.

\paragraph{MLP (7.4k parameters).} Three hidden ReLU layers of width 58 on the
flattened input. Unconstrained reference.

\paragraph{MLP-Aug.} The same MLP trained with 8x random SO(3) data
augmentation, rotating all input vectors and the target with the same
rotation.

\paragraph{Scalarization (1.3k to 1.5k parameters).} All pairwise dot products
of the $k$ input vectors feed a two-layer MLP of width 32 that outputs one
coefficient per basis element $\{v_i\} \cup \{v_i \times v_j\}_{i<j}$; the
output is the coefficient-weighted sum. Exactly SO(3)-equivariant. For $k=3$
a strengthened variant adds the scalar triple product
$v_1 \cdot (v_2 \times v_3)$ to the invariants, making orientation-sensitive
information directly available to the coefficient network.

\paragraph{Vector Neurons (2.4k to 2.6k parameters).} An external equivariant
baseline~\citep{deng2021vector}: channels are 3D vectors, linear layers mix
channels with learned scalars, and the VN-ReLU nonlinearity projects each
feature on the halfspace defined by a learned direction. Faithful to the
original design, the output lies in the linear span of the input channels,
so a two-vector input cannot produce a cross product. We therefore also test
VN-Cross, the common practical fix that appends pairwise cross products of
the inputs as extra channels.

\paragraph{E3NN (6.6k to 11k parameters).} A second external baseline built
from irreducible-representation tensor products, the standard construction
behind irreps equivariant models~\citep{thomas2018tensor,geiger2022e3nn}.
Inputs are $l{=}1$ irreps, each block is a self tensor product gated by scalar
channels with an equivariant BatchNorm, and the norm of each input vector is
expanded into a small Gaussian radial basis fed as extra scalar channels. This
radial embedding is standard in irreps models~\citep{batzner2022nequip} and
was necessary here: without it the network cannot represent functions of an
angle magnitude and stays worse than a plain MLP on the rotation task. As this
benchmark is SO(3) rather than full O(3), all irreps use a single parity, which
sidesteps a channel-pruning failure mode we observed when mixing parities.

\paragraph{GeoBilinear (28k parameters).} Inputs are embedded as multivectors
of $\mathrm{Cl}(3,0)$ (8 components per channel: vectors on grade 1, axis-angle
inputs on grade 2 via duality). Three stacked blocks compute
$\mathrm{gp}(Ax, Bx) + Cx$ where $\mathrm{gp}$ is the geometric product and
$A, B, C$ are unconstrained linear maps on multivector channels, followed by a
grade-norm gate and channel normalization. Ablation control: geometric
products without the equivariant constraint.

\paragraph{GeoEquivariant (2k parameters).} The same architecture with
grade-wise weight tying (one scalar per grade per channel pair), which makes
every layer exactly SO(3)-equivariant. Measured relative equivariance error is
$2\times 10^{-7}$, the float32 limit. The augmented MLP, for comparison,
reaches 0.22 to 0.49 depending on the task.

\section{Tasks and protocol}
\label{sec:tasks}

Inputs are standard normal unless stated. Train and test sets use disjoint
random generator seeds; out-of-distribution (OOD) splits use disjoint input
regions, verified numerically. Five seeds per configuration; we report mean
normalized MSE (NMSE): test MSE divided by the MSE of the constant mean
predictor on the test targets. NMSE 1.0 equals a trivial predictor. We use
NMSE because raw MSE is misleading under distribution shift: on far-radius
splits the targets are small, so a failing model still gets a small raw MSE.

\begin{enumerate}
\item \textbf{rotation}: $(u, p) \mapsto R(u)\,p$ with $u$ an axis-angle
vector. OOD: unseen axis hemisphere; unseen angle range $[\pi/2, \pi]$ after
training on $[0, \pi/2]$.
\item \textbf{cross}: $(a, b) \mapsto a \times b$. OOD: unseen hemisphere.
\item \textbf{central force}: $(r, d) \mapsto -r/(\lVert r\rVert^3 + 0.05)$
with a distractor input $d$. OOD: radii $[1.25, 3]$ after training on
$[0.25, 1]$.
\item \textbf{two-body force}: $(r_1, r_2) \mapsto
-(r_1 - r_2)/(\lVert r_1 - r_2\rVert^3 + 0.05)$. OOD: separations $[2, 4]$
after training on $[0.5, 1.5]$.
\item \textbf{composed rotations}: $(u_1, u_2, p) \mapsto R(u_2) R(u_1)\, p$.
OOD: unseen hemisphere for both axes.
\item \textbf{local-to-world torque}: $(r, u, f) \mapsto r \times (R(u) f)$,
a local force mapped to the world frame and applied at a lever arm, the
minimal robotics composition. OOD: unseen axis hemisphere; unseen angle range.
\end{enumerate}

We call tasks 1 to 4 single-stage: their targets require one equivariant
interaction of the inputs, possibly modulated by a learned function of
invariants. Tasks 5 and 6 are compositional: their computation graph nests
group operations. All tasks lie within the intended approximation class of
both equivariant model families, and neither family faces a symmetry-induced
expressivity obstruction on these targets, so differences measure
optimization behavior and inductive fit rather than hard expressivity gaps.

\section{Results}
\label{sec:results}

\begin{figure}[t]
\centering
\includegraphics[width=0.85\linewidth]{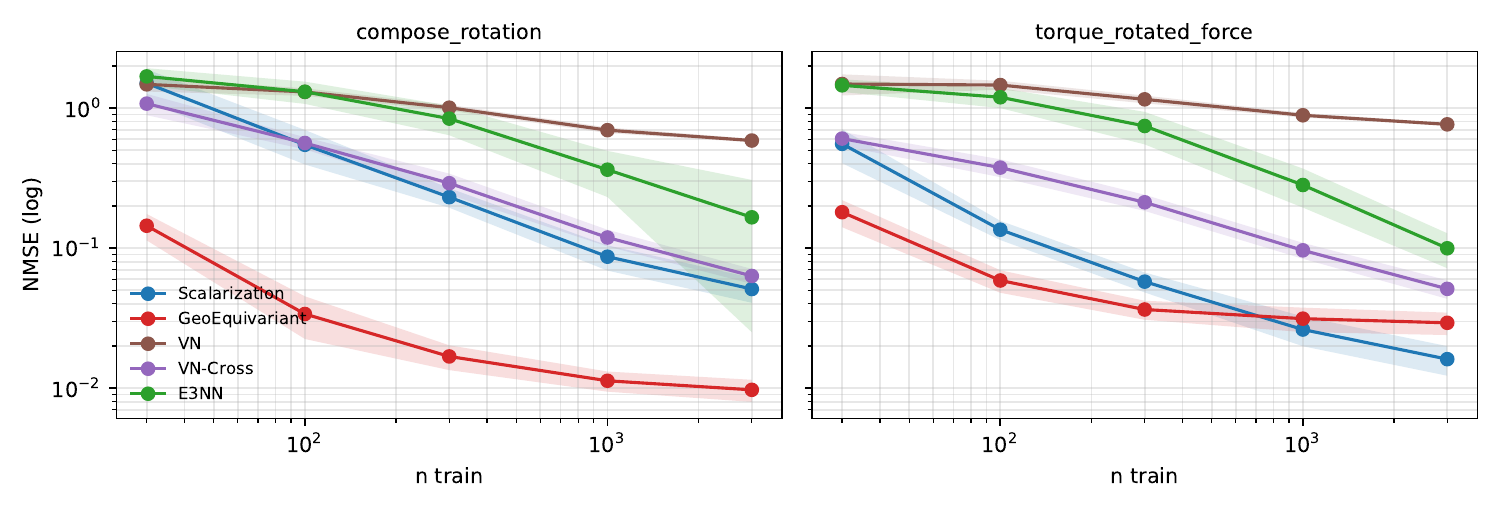}
\caption{Sample efficiency on the compositional tasks, NMSE versus training
set size, mean over 10 seeds, shaded bands are one standard deviation. On
composed rotations the geometric network at $n{=}100$ already beats
scalarization at $n{=}3000$ and the gap never closes. On the torque task the
advantage is confined to low data: scalarization overtakes beyond
$n \approx 1000$.}
\label{fig:curves}
\end{figure}

\begin{figure}[t]
\centering
\includegraphics[width=0.95\linewidth]{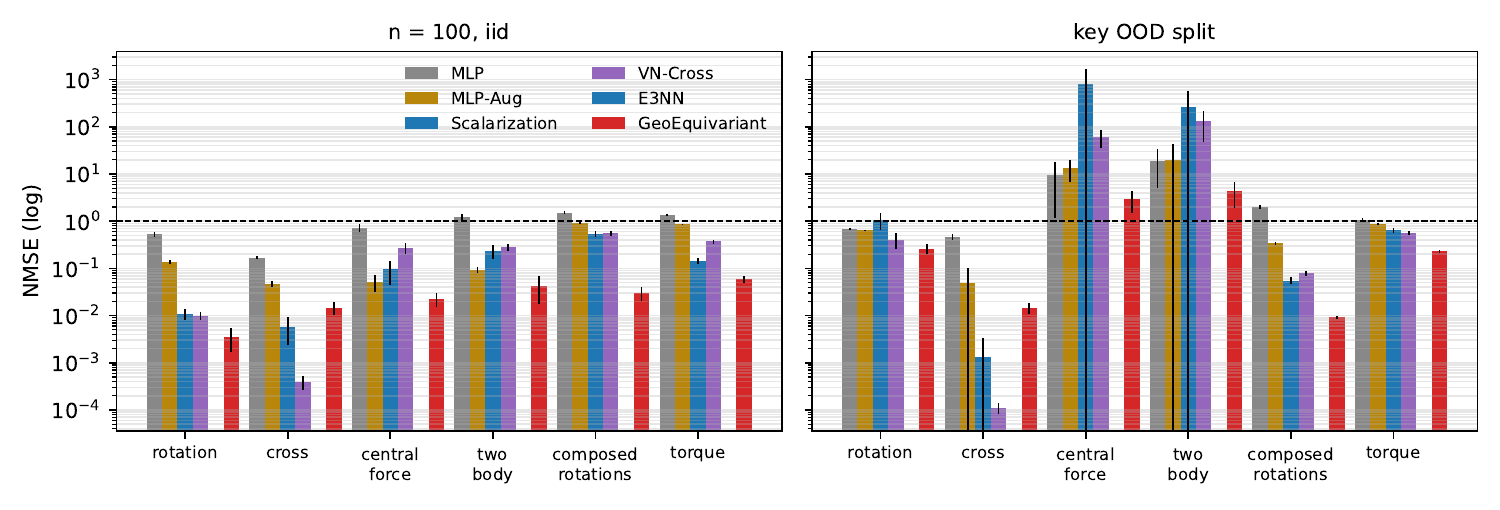}
\caption{NMSE overview per task and model (log scale, mean over 5 seeds,
error bars over seeds). Right panel OOD splits: angle shift for rotation and
torque, axis shift for cross and composed rotations, radius shift for central
force, separation shift for two-body. The dashed line marks the constant
predictor. Generated by \texttt{paper/make\_figures.py}.}
\label{fig:overview}
\end{figure}

\subsection{Finding 1: exact equivariance dominates, whatever its construction}

At $n{=}100$ on the rotation task, NMSE is 0.53 for the MLP, 0.14 for the
augmented MLP, 0.011 for scalarization and 0.0036 for GeoEquivariant.
Augmentation reduces the measured equivariance error (0.28 to 0.22) but stays
10 to 20x behind the exact constructions on directional OOD splits. This
pattern holds on all six tasks (Table~\ref{tab:simple},
Table~\ref{tab:compositional}).

\subsection{Finding 2: on single-stage laws, scalarization is enough}

\begin{table}[t]
\centering
\caption{Single-stage tasks, NMSE (mean $\pm$ std over 5 seeds), 200 epochs.
The OOD split is the axis shift for rotation and cross, the radius shift for
central force, and the separation shift for two-body. Values above 1.0 are
worse than a constant predictor; the large OOD entries are real failures,
amplified by normalization against small far-field targets.
\texttt{s/run} is mean training wall-clock per run (CPU).}
\label{tab:simple}
\scriptsize
\begin{tabular}{llrrrrr}
\toprule
Task & Model & Params & $n{=}100$ iid & $n{=}1000$ iid & OOD & s/run \\
\midrule
rotation & MLP & 7427 & $0.53 \pm 0.06$ & $0.101 \pm 0.006$ & $0.61 \pm 0.07$ & 1.0 \\
rotation & MLP-Aug & 7427 & $0.14 \pm 0.01$ & $0.054 \pm 0.005$ & $0.064 \pm 0.007$ & 3.9 \\
rotation & Scalarization & 1283 & $0.011 \pm 0.003$ & $0.0040 \pm 0.0016$ & $0.0033 \pm 0.0011$ & 0.6 \\
rotation & VN & 2424 & $0.46 \pm 0.02$ & $0.41 \pm 0.01$ & $0.41 \pm 0.01$ & 6.8 \\
rotation & VN-Cross & 2472 & $0.010 \pm 0.002$ & $0.0029 \pm 0.0007$ & $0.0026 \pm 0.0003$ & 7.5 \\
rotation & GeoBilinear & 28352 & $0.59 \pm 0.07$ & $0.059 \pm 0.004$ & $0.75 \pm 0.05$ & 16.9 \\
rotation & E3NN & 6612 & $0.36 \pm 0.10$ & $0.048 \pm 0.019$ & $0.031 \pm 0.011$ & 38.8 \\
rotation & GeoEquivariant & 1952 & $0.0036 \pm 0.0019$ & $0.0031 \pm 0.0008$ & $0.0031 \pm 0.0010$ & 18.7 \\
\midrule
cross & MLP & 7427 & $0.17 \pm 0.01$ & $0.040 \pm 0.011$ & $0.46 \pm 0.06$ & 0.6 \\
cross & MLP-Aug & 7427 & $0.047 \pm 0.007$ & $0.025 \pm 0.016$ & $0.049 \pm 0.051$ & 2.0 \\
cross & Scalarization & 1283 & $0.0058 \pm 0.0034$ & $0.00054 \pm 0.00047$ & $0.0013 \pm 0.0020$ & 0.7 \\
cross & VN & 2424 & $1.000 \pm 0.000$ & $1.000 \pm 0.000$ & $1.000 \pm 0.001$ & 4.7 \\
cross & VN-Cross & 2472 & $0.00039 \pm 0.00013$ & $0.00012 \pm 0.00002$ & $0.00011 \pm 0.00003$ & 4.7 \\
cross & GeoBilinear & 28352 & $0.58 \pm 0.14$ & $0.025 \pm 0.002$ & $1.01 \pm 0.22$ & 12.8 \\
cross & E3NN & 6612 & $0.44 \pm 0.09$ & $0.056 \pm 0.012$ & $0.075 \pm 0.023$ & 14.9 \\
cross & GeoEquivariant & 1952 & $0.015 \pm 0.005$ & $0.014 \pm 0.003$ & $0.014 \pm 0.004$ & 10.8 \\
\midrule
central force & MLP & 7427 & $0.73 \pm 0.14$ & $0.027 \pm 0.006$ & $9.6 \pm 8.5$ & 1.0 \\
central force & MLP-Aug & 7427 & $0.052 \pm 0.020$ & $0.0042 \pm 0.0008$ & $13.4 \pm 6.6$ & 2.6 \\
central force & Scalarization & 1283 & $0.095 \pm 0.050$ & $0.031 \pm 0.008$ & $817 \pm 878$ & 0.6 \\
central force & VN & 2424 & $0.73 \pm 0.16$ & $0.48 \pm 0.02$ & $150 \pm 14$ & 4.6 \\
central force & VN-Cross & 2472 & $0.27 \pm 0.07$ & $0.13 \pm 0.05$ & $60 \pm 24$ & 3.8 \\
central force & GeoBilinear & 28352 & $1.48 \pm 0.19$ & $0.063 \pm 0.005$ & $17.8 \pm 2.7$ & 11.8 \\
central force & E3NN & 6612 & $0.84 \pm 0.47$ & $0.21 \pm 0.15$ & $43 \pm 6$ & 22.2 \\
central force & GeoEquivariant & 1952 & $0.023 \pm 0.007$ & $0.0073 \pm 0.0015$ & $2.9 \pm 1.4$ & 12.2 \\
\midrule
two-body & MLP & 7427 & $1.24 \pm 0.21$ & $0.036 \pm 0.009$ & $19.2 \pm 14.2$ & 0.7 \\
two-body & MLP-Aug & 7427 & $0.093 \pm 0.013$ & $0.0057 \pm 0.0010$ & $19.8 \pm 22.5$ & 2.3 \\
two-body & Scalarization & 1283 & $0.23 \pm 0.07$ & $0.11 \pm 0.03$ & $268 \pm 291$ & 0.6 \\
two-body & VN & 2424 & $0.93 \pm 0.26$ & $0.49 \pm 0.04$ & $151 \pm 12$ & 4.9 \\
two-body & VN-Cross & 2472 & $0.28 \pm 0.05$ & $0.12 \pm 0.02$ & $134 \pm 86$ & 7.1 \\
two-body & GeoBilinear & 28352 & $1.77 \pm 0.28$ & $0.11 \pm 0.02$ & $9.8 \pm 1.9$ & 13.1 \\
two-body & E3NN & 6612 & $1.9 \pm 1.8$ & $0.40 \pm 0.39$ & $75 \pm 32$ & 15.7 \\
two-body & GeoEquivariant & 1952 & $0.042 \pm 0.025$ & $0.0064 \pm 0.0028$ & $4.4 \pm 2.5$ & 13.2 \\
\bottomrule
\end{tabular}
\end{table}

On the cross product at $n{=}1000$, scalarization reaches NMSE
$5.4\times 10^{-4}$ against $1.4\times 10^{-2}$ for GeoEquivariant, a 26x gap.
With a modest tuning grid, scalarization also beats GeoEquivariant on rotation
($4.7\times 10^{-4}$ vs $3.1\times 10^{-3}$) and central force
($1.2\times 10^{-3}$ vs $7.3\times 10^{-3}$). GeoEquivariant plateaus around
NMSE $3\times 10^{-3}$ to $1.4\times 10^{-2}$ on these tasks. The ablation
explains the plateau: removing EquiNorm lets the best seeds reach
$9\times 10^{-5}$ on the cross product, matching the strongest baselines, but
training becomes unreliable (seed range $10^{-4}$ to $0.1$, and up to 1.1 on
rotation). The channel normalization trades peak fit for optimization
stability; GradeGate removal changes little. The external
baselines sharpen the picture: vanilla Vector Neurons sit exactly at the
trivial predictor on the cross product (NMSE 1.00, the target leaves the
linear span of the inputs), while VN-Cross reaches $1.2\times 10^{-4}$ at
$n{=}1000$, the best result of any model on that task. The irreps baseline
E3NN, despite being the standard construction of the field, is the weakest of
the three equivariant models on every single-stage task (cross at $n{=}100$:
0.44, against 0.006 for scalarization), which underscores that generic
equivariance without a task-matched parameterization is not enough at this
scale. Any claim that $\mathrm{Cl}(3,0)$ layers as such help on simple vector
laws is not supported.

\subsection{Finding 3: composition is where geometric algebra pays}

\begin{table}[t]
\centering
\caption{Compositional tasks, NMSE (mean over 5 seeds), 200 epochs.}
\label{tab:compositional}
\small
\begin{tabular}{llrrrrr}
\toprule
Task & Model & Params & $n{=}100$ iid & $n{=}1000$ iid & OOD axis & OOD angle \\
\midrule
composed rotations & MLP & 7601 & 1.51 & 0.68 & 2.02 & n/a \\
composed rotations & MLP-Aug & 7601 & 0.92 & 0.30 & 0.34 & n/a \\
composed rotations & Scalarization & 1478 & 0.54 & 0.087 & 0.055 & n/a \\
composed rotations & GeoBilinear & 29888 & 1.23 & 0.54 & 1.04 & n/a \\
composed rotations & VN & 2472 & 1.28 & 0.67 & 0.61 & n/a \\
composed rotations & VN-Cross & 2616 & 0.56 & 0.12 & 0.079 & n/a \\
composed rotations & E3NN & 11022 & 1.35 & 0.41 & 0.19 & n/a \\
composed rotations & GeoEquivariant & 2048 & \textbf{0.031} & \textbf{0.012} & \textbf{0.0092} & n/a \\
\midrule
torque & MLP & 7601 & 1.35 & 0.66 & 1.16 & 1.08 \\
torque & MLP-Aug & 7601 & 0.85 & 0.37 & 0.40 & 0.87 \\
torque & Scalarization & 1478 & 0.145 & \textbf{0.029} & \textbf{0.020} & 0.64 \\
torque & GeoBilinear & 29888 & 1.63 & 0.58 & 0.66 & 0.56 \\
torque & VN & 2472 & 1.52 & 0.88 & 0.79 & 0.80 \\
torque & VN-Cross & 2616 & 0.37 & 0.090 & 0.064 & 0.56 \\
torque & E3NN & 11022 & 1.13 & 0.31 & 0.15 & 0.88 \\
torque & GeoEquivariant & 2048 & \textbf{0.059} & 0.030 & 0.031 & \textbf{0.23} \\
\bottomrule
\end{tabular}
\end{table}

\begin{table}[t]
\centering
\caption{Strengthened scalarization control on the compositional tasks
(5 seeds). Widening the baseline and adding the triple-product invariant does
not close the gap.}
\label{tab:control}
\small
\begin{tabular}{llrrrr}
\toprule
Task & Variant & Params & $n{=}100$ & $n{=}1000$ & key OOD \\
\midrule
composed rot. & scalar.\ h32 & 1478 & 0.54 & 0.087 & 0.055 (axis) \\
composed rot. & scalar.\ h128 d3 + triple & 34822 & 0.29 & 0.038 & 0.020 (axis) \\
composed rot. & GeoEquivariant & 2048 & \textbf{0.031} & \textbf{0.012} & \textbf{0.0092} (axis) \\
\midrule
torque & scalar.\ h128 & 18182 & 0.139 & 0.028 & 0.46 (angle) \\
torque & scalar.\ h128 d3 + triple & 34822 & 0.146 & \textbf{0.020} & 0.43 (angle) \\
torque & GeoEquivariant & 2048 & \textbf{0.059} & 0.030 & \textbf{0.23} (angle) \\
\bottomrule
\end{tabular}
\end{table}

On composed rotations, GeoEquivariant reaches NMSE 0.031 / 0.012 / 0.0092
($n{=}100$ / $n{=}1000$ / OOD axes) against 0.54 / 0.087 / 0.055 for
scalarization. Strengthening the baseline with the triple-product invariant,
width 128 and depth 3 (35k parameters against 2k) only closes the gap to
0.29 / 0.038 / 0.020, still 9x / 3x / 2x behind (Table~\ref{tab:control}).
The result holds against both external baselines: on composed rotations
VN-Cross stays 6x to 17x behind GeoEquivariant at every training size
(0.12 vs 0.012 at $n{=}1000$), vanilla Vector Neurons never leave the
0.6 to 1.3 range, and E3NN, though a native irreps architecture, is worse
still in low data (1.35 at $n{=}100$, over 40x behind) and only reaches 0.19 on
the OOD axis split against 0.0092. The sample-efficiency curves (Figure~\ref{fig:curves},
10 seeds) sharpen this: on composed rotations GeoEquivariant at $n{=}100$
(NMSE 0.034) already beats scalarization at $n{=}3000$ (0.051), the gap is
16x at $n{=}100$ and still 5x at $n{=}3000$. The pattern replicates on the torque task with an
honest caveat: the advantage concentrates in low data (3x at $n{=}30$) and
under angle shift (2.8x against the plain baseline, 1.9x against the
strongest variant), but scalarization crosses over near $n{=}1000$ and leads
at $n{=}3000$ (0.016 vs 0.029). The unconstrained MLP fails on the torque
task even iid at $n{=}1000$ (NMSE 0.66).

The mechanism: a rotor is an element of $\mathrm{Cl}(3,0)$ and rotor
composition is one geometric product, so depth-stacked products represent the
target natively, while a fixed equivariant basis with learned invariant
coefficients must approximate the composition map through its scalar MLP.
Three ablations test this reading directly (Table~\ref{tab:depth}).

First, on chains of $L$ rotations ($p \mapsto R_L \cdots R_1 p$,
$n{=}1000$), the network depth required for low error tracks the composition
depth: each added block gains roughly an order of magnitude on longer chains,
and depth 1 fails even at $L{=}1$ (a rotor sandwich takes two products).
Scalarization degrades catastrophically with chain length, from 0.004 at
$L{=}1$ to worse than the constant predictor at $L{=}4$ (NMSE 1.9), while the
depth-4 geometric network stays at 0.062.

Second, replacing the ReLU coefficient network of scalarization by
multiplicative units, $(Ax) \odot (Bx) + Cx$, does not close the gap
(compose\_rotation $n{=}1000$: 0.22 against 0.012) and is much worse in low
data (7.1 at $n{=}100$). Generic multiplicative depth is not the explanation;
the typed Clifford product is doing specific work.

Third, composition alone is not where the advantage comes from. On the
rotation-free composition $(a,b,c) \mapsto a \times (b \times c)$,
scalarization wins by 24x over GeoEquivariant (0.0028 against 0.068 at
$n{=}1000$). The reason is classical: the identity
$a \times (b \times c) = b\,(a \cdot c) - c\,(a \cdot b)$ flattens this
target into polynomial invariant coefficients on the fixed basis, exactly
scalarization's regime. Chained rotations also flatten in principle, but into
increasingly complex coefficient functions of the invariants. The refined
claim is therefore: the geometric advantage tracks the complexity of the
invariant-coefficient functions needed to flatten the target, not
composition per se.

\begin{table}[t]
\centering
\caption{Depth against composition depth: NMSE (mean over 5 seeds) of
GeoEquivariant with depth 1 to 4 on chains of $L$ rotations, $n{=}1000$ iid,
with scalarization as reference. Bold: best per row.}
\label{tab:depth}
\small
\begin{tabular}{lrrrrr}
\toprule
Chain & depth 1 & depth 2 & depth 3 & depth 4 & Scalarization \\
\midrule
$L{=}1$ & 0.42 & 0.013 & 0.0031 & \textbf{0.0019} & 0.0040 \\
$L{=}2$ & 0.64 & 0.14 & 0.012 & \textbf{0.0074} & 0.090 \\
$L{=}3$ & 0.77 & 0.30 & 0.048 & \textbf{0.027} & 0.74 \\
$L{=}4$ & 0.87 & 0.50 & 0.15 & \textbf{0.062} & 1.91 \\
\bottomrule
\end{tabular}
\end{table}

\subsection{Finding 4: nobody extrapolates invariant magnitudes}

On angle, radius and separation shifts, every model degrades badly. On the
radial splits all models are worse than the constant predictor:
GeoEquivariant reaches NMSE 2.9 (central force, far radii) and 4.4 (two-body,
far separations); the MLP reaches 9.6 and 19; scalarization degrades worst,
up to 817, because its learned radial coefficients extrapolate freely.
Equivariance constrains directions, not the response to invariant magnitudes.
On angle shift the equivariant models remain below trivial but far from
solved (0.23 to 0.26); GeoEquivariant only degrades more slowly than
scalarization (0.43 to 1.8). We note that unnormalized MSE inverts this
conclusion on the radial splits, because far-field targets are small; earlier
internal reports of this project made exactly that mistake.

\subsection{Cost}

GeoEquivariant is the slowest model per run at these sizes, 13 to 19x the MLP
and scalarization training time, despite having the fewest parameters among
the learned models except scalarization. Scalarization is the cheapest model
across the board.

\section{Controls}
\label{sec:controls}

\paragraph{Optimization fairness.} All models in the main tables received an
identical fixed budget: Adam, full batch, 200 epochs, learning rate
$5\times 10^{-3}$, one training run per seed, no early stopping, no model
selection. A tuning grid over learning rate
$\{10^{-3}, 5\times 10^{-3}, 2\times 10^{-2}\}$ and epochs $\{500, 2000\}$ at
$n{=}1000$ was additionally applied to the two baselines most at risk of
being under-tuned: it improves the MLP by 30 to 50 percent (rotation: 0.101
to 0.072) and improves scalarization to the values quoted in Finding 2.
GeoEquivariant and the Vector Neurons variants were not tuned, so the grid
can only favor the baselines. Wall-clock numbers are means over all runs of a
configuration on an Intel i5-5300U (4 threads, PyTorch 2.12 CPU); the
sample-efficiency curves ran on an RTX 4070 Ti with identical results on
overlapping configurations.

\paragraph{Seed stability.} Per-seed values for every cell are in the
repository. The large OOD means are heavy-tailed across seeds; medians tell
the same story (central force OOD radius: scalarization 546, MLP 7.3,
GeoEquivariant 2.3; two-body OOD separation: GeoEquivariant 3.0), so no
conclusion rests on an outlier seed.

\paragraph{Data and split hygiene.} Train and test generators are disjoint;
OOD regions are disjoint by construction and verified numerically (hemisphere
sign fractions, radius ranges).\footnote{A task added during exploration
(\texttt{local\_force}) turned out to be the rotation task under a different
name, with per-seed identical results. It is excluded from the evidence and
documented in the repository.}

\section{Limitations}
\label{sec:limitations}

Synthetic tasks with exactly realizable targets; two compositional tasks,
both built from the same rotation primitive. The external baselines (Vector
Neurons, e3nn) are compact untuned instances, not the large tuned models these
frameworks reach on real datasets, so their weakness here speaks to the
low-data small-scale regime, not to the frameworks in general. Five seeds on
the main tables. Results are small scale (thousands of samples, 2k-parameter
models); we make no claim about behavior at scale. The flattening-complexity
reading of Finding 3 is supported by the ablations but not yet formalized.

\section{Conclusion}

Exact SO(3) equivariance, not geometric algebra, explains most of the gains
that small $\mathrm{Cl}(3,0)$ networks show on simple 3D vector laws: a
trivial scalarization baseline matches or beats them at a fraction of the
cost. The geometric product earns its place when the target composes group
elements, where it wins by 2.5x to 9x in low data and under angle shift
against strengthened baselines. Neither approach extrapolates invariant
magnitudes, and unnormalized metrics can hide this failure. For practitioners
the guidance is: reach for scalarization first, including for compositions
that flatten into simple invariant coefficients; consider geometric algebra
layers when the law chains group operations whose flattened coefficient
functions are complex, especially in low data, keeping in mind that the
advantage is task-dependent and can disappear with more data.

\bibliographystyle{plainnat}
\bibliography{references}

\end{document}